\RequirePackage{amsthm}

\documentclass{article}

\usepackage{bm}
\usepackage{orcidlink}
\usepackage{comment}
\usepackage{csquotes}
\usepackage{makecell}
\usepackage{tabularx}
\usepackage[table,xcdraw]{xcolor}
\usepackage{anyfontsize}
\usepackage{graphicx}%
\usepackage{multirow}%
\usepackage{amsmath,amssymb,amsfonts}%
\usepackage{amsthm}%
\usepackage{mathrsfs}%
\usepackage[title]{appendix}%
\usepackage{textcomp}%
\usepackage{manyfoot}%
\usepackage{booktabs}%
\usepackage{algorithm}%
\usepackage{algorithmicx}%
\usepackage{algpseudocode}%
\usepackage{listings}%
\usepackage{natbib}
\usepackage{authblk}
\usepackage{orcidlink}
\usepackage{fancyhdr}

\fancypagestyle{firstpage}{
  \fancyhf{}
  \fancyhead[C]{%
      \fcolorbox{black}{gray!10}{%
          \parbox{0.9\textwidth}{%
              \small
              \textbf{Preprint notice.}
              This paper is a preliminary version of an accepted article in
              \emph{Information Systems Frontiers}, Springer, Special Issue
              \emph{“Explainability in Human-Centric AI”}. Please cite the final published version of the paper, not this preprint. The final published version can be found at \url{https://link.springer.com/article/10.1007/s10796-025-10682-3} (doi:10.1007/s10796-025-10682-3).
}
  }}
  \setlength{\headheight}{2pt}
  \setlength{\headsep}{45pt}
}

\newcommand{\abs}[1]{\lvert#1\rvert}

\theoremstyle{thmstyleone}%
\theoremstyle{thmstyletwo}%
\theoremstyle{thmstylethree}%
\begin{document}

\title{\textbf{Trustworthy Data-driven Chronological Age Estimation from Panoramic Dental Images}}

\author[1,2]{Ainhoa Vivel-Couso\orcidlink{0000-0002-5860-4849}\thanks{Corresponding author: ainhoa.vivel.couso@usc.es \par Contributing authors: nicolas.vila@usc.es; mariajose.carreira@usc.es;
\mbox{alberto.bugarin.diz@usc.es; inmaculada.tomas@usc.es; josemaria.alonso.moral@usc.es}}}
\author[1,2,4]{Nicolás Vila-Blanco\orcidlink{0000-0001-5865-9973}}
\author[1,2,4]{María J. Carreira\orcidlink{0000-0003-0532-2351}}
\author[1,2]{Alberto Bugarín-Diz\orcidlink{0000-0003-3574-3843}}
\author[1,3,4]{Inmaculada Tomás\orcidlink{0000-0002-3317-0853}}
\author[1,2]{Jose M. Alonso-Moral\orcidlink{0000-0003-3673-421X}}

\affil[1]{Centro Singular de Investigación en Tecnoloxías Intelixentes (CiTIUS), Universidade de Santiago de Compostela, Santiago de Compostela, Spain}
\affil[2]{Department of Electronics and Computing, Universidade de Santiago de Compostela, Santiago de Compostela, Spain}
\affil[3]{Instituto de Investigación Sanitaria de Santiago de Compostela (IDIS), Universidade de Santiago de Compostela, Santiago de Compostela, Spain}
\affil[4]{Oral Sciences Research Group, Universidade de Santiago de Compostela, Santiago de Compostela, Spain}

\date{}
\maketitle
\thispagestyle{firstpage}

\vspace{-2em}

\begin{abstract}
Integrating deep learning into healthcare enables personalized care but raises trust issues due to model opacity. To improve transparency, we propose a system for dental age estimation from panoramic images that combines an opaque and a transparent method within a natural language generation (NLG) module. This module produces clinician-friendly textual explanations about the age estimations, designed with dental experts through a rule-based approach. Following the best practices in the field, the quality of the generated explanations was manually validated by dental experts using a questionnaire. The results showed a strong performance, since the experts rated 4.77$\pm$0.12 (out of 5) on average across the five dimensions considered. We also performed a trustworthy self-assessment procedure following the ALTAI checklist, in which it scored 4.40$\pm$0.27 (out of 5) across seven dimensions of the AI Trustworthiness Assessment List.
\end{abstract}

\noindent\textbf{Keywords:} Human-Centric Explainable Artificial Intelligence, Data-to-Text Systems, Ruled-based Text Generation, Fuzzy Quantification, Surrogate Deep Learning

\section{Introduction}\label{sec1}
The use of Artificial Intelligence (AI) in medicine and healthcare faces important risks (e.g., lack of transparency, privacy and security, gaps in accountability, etc.)~\citep{Lekadir2022}. Therefore, there is international consensus that trustworthy and deployable AI in healthcare should be fair, universal, traceable, usable, robust, and explainable~\citep{Lekadire081554}. 

Nowadays, integrating machine learning algorithms into medicine has precipitated a revolutionary transformation in healthcare practices, particularly in the medical imaging domain, where AI models can discern intricate patterns, formulate accurate predictions, and furnish invaluable insights that were previously inaccessible \citep{bioengineering10121435}. Machine learning can potentially enhance outcomes and optimize healthcare delivery by facilitating disease detection, optimizing treatment strategies, and personalizing patient care \citep{haug2023artificial}. By enabling data-driven decision-making and paving the way for precision medicine, the revolution machine learning brings to medicine holds great promise for transforming how we approach healthcare~\citep{doi:10.1177/0022034520915714}.

The specific application of deep learning has led to a paradigm shift in the effective utilization of medical data. The availability of extensive medical image datasets and advancements in computational resources have enabled deep learning to become a powerful tool in the field. Its capacity to process intricate image representations and discern complex patterns with minimal oversight has led to significant advancements in numerous medical procedures \citep{chen2022recent}.

Although integrating machine learning (particularly deep learning) in medical imaging has immense potential, there is an urgent need for computer-assisted support systems to be easily comprehensible to medical professionals \citep{li2022interpretable}. As healthcare decisions are contingent upon the expertise of medical professionals, it is imperative to possess a deep understanding of the mechanisms underlying such systems' predictions. Interpretability is paramount in ensuring transparency and empowering clinicians to validate system outputs, discern potential biases or errors, and formulate informed decisions based on a comprehensive understanding of the underlying reasoning. 

In this regard, deep learning models are characterized by a lack of interpretability \citep{hayashi2020black}, its opaque nature and inherent complexity, which poses significant challenges in elucidating the rationale behind specific decisions~\citep{guidotti_survey_2018, adadi2018}. Furthermore, these models may unwittingly learn and perpetuate existing biases in the training data, resulting in biased predictions and inequitable healthcare outcomes. This potential for bias in deep learning models can hinder their acceptance and trustworthiness in clinical practice. Addressing and mitigating these biases is critical to ensure equitable and fair decision-making processes~\citep{ANTAMIS2024128204, li2022interpretable}.

The scientific community is developing techniques that enhance trustworthiness in deep learning models to overcome these challenges. Specifically, research in Explainable AI (XAI) seeks to elucidate the decision-making process of these models and provide medical professionals with meaningful explanations for their predictions~\citep{ali2023explainable}. This includes techniques closely related to computer vision, such as saliency maps~\citep{JIN2023102009,vila2020deep} or gradient-based methods~\citep{morrison2023shared}, which aim to highlight the relevant parts of the images that support the predictions.

Furthermore, Natural Language Generation (NLG) has emerged as a promising approach to enhance the interpretability of medical imaging algorithms based on deep learning \citep{bookReiter2024}, since NLG systems can automatically generate human-readable descriptions and explanations of the algorithms' output \citep{messina2022survey}. Converting complex, often abstract image-based predictions into comprehensible textual descriptions enables clinicians to discern the rationale behind the algorithms, validate their behavior, and ultimately make informed decisions.

Moreover, the EU AI Act~\citep{EURegulation1689_2024} entered into force on 1 August 2024.
This risk-based regulation fosters responsible AI development and deployment, with healthcare AI systems classified as high-risk systems. Consequently, these systems are subject to stringent regulatory requirements, necessitating heightened transparency, straightforward design, and robust validation, among other factors. 

In this work, we introduce AgeX, a human-centric intelligent system that generates textual explanations of the predictions made by a non-transparent deep learning model to estimate a patient's age and sex based on dental data. Our primary focus was chronological age estimation from panoramic dental images (orthopantomogram, OPG), a crucial aspect in various clinical procedures. These include body identification, legal age determination in criminal proceedings, migration control, and adoption processes, among others \citep{vandenberghe2010modern}. The central motivation of this research is to demonstrate the feasibility of a transparent and interpretable method for explaining the results generated by an automatic and accurate, but opaque model. We have followed a trustworthiness by design approach, considering requirements such as robustness, interpretability and explainability that were deemed essential by the odontologist included in the development team. Notice that the use of an interpretable method in this context is particularly important, particularly in scenarios where the deep learning methods employed are non-traceable, thereby providing a level of interpretability that would not otherwise be acceptable. 
Therefore, our work is designed under the provisions outlined in the EU AI Act, leveraging automated natural language descriptions to enhance transparency. AgeX also complies with the transparency requirements stipulated by the European AI Act, since it informs end-users about which content was AI-generated. Additionally, the objective is to ascertain the reliability of this trustworthy, automatic, high-performance approach for deployment in a realistic age estimation scenario which was evaluated by a panel of expert dental professionals. 

The remainder of the manuscript is structured as follows. 
Section~\ref{sota} is a comprehensive review of current advancements in the field, which establishes the foundation for understanding the importance of this study. Section~\ref{prop} introduces our approach to trustworthy decision-making support, which combines transparent and opaque methods for predicting dental age alongside a novel method to explain predictions in natural language. 
Section~\ref{exp} delineates the experimental framework, investigates the correlation between the transparent and opaque predictive methods, and delves into the intricacies of the evaluation process.
Section~\ref{cap:test} presents a comprehensive statistical analysis of dental attributes and correlations between opaque and transparent methods. It also addresses data processing, discusses the generation and evaluation of predictions and related explanations, and concludes with a discussion on ethical considerations. Finally, Section~\ref{conc} summarizes the findings and outlines potential directions for further research, highlighting the implications of our work for the field of chronological age estimation.

\section{State of the Art}
\label{sota}

\subsection{Dental-Based Chronological Age Estimation}
Among all the body parts used to estimate the (unknown) age of an individual, the dentition stands out as the most commonly used structure, mainly because tooth development patterns are highly correlated with chronological age and are less affected by external factors~\citep{willems2001review}. Consequently, dental radiographs have emerged as a highly effective and non-destructive method for these estimations. 

The methods proposed in the literature vary depending on the specific dental indicator used to predict the chronological age, with most of them focused on individuals with developing teeth. 
The initial approaches relied on the clinical eruption as the primary indicator, given its ease of observation~\citep{haavikko1970formation}. However, the eruption process is relatively brief. Consequently, its application is constrained to a relatively limited range of individuals. Accordingly, researchers have developed atlases that relate different dental events (in addition to clinical eruption) to chronological age for use in a wider population cohort~\cite {willems2001review}. The most prominent example is the London Atlas~\citep{alqahtani2010brief}. 
The subsequent pivotal development was the introduction of scoring systems. These systems proposed a division of the dental development of each tooth into several stages and substages, where each stage corresponds to a specific numeric score. Through the application of these systems, researchers were able to translate the combination of scores for each tooth into a numeric chronological age, facilitating a more systematic analysis of dental development. The most widely adopted methodologies in this regard were proposed by~\cite{nolla1960development} with 11 stages and~\cite{demirjian1973new} with 8 stages. 
Additionally, \cite{cameriere2006age}, demonstrated a positive correlation between the openness of the dental apices and chronological age. Consequently, they developed a formula to estimate the dental age considering only permanent teeth.

The advent of computer image processing has had a profound impact on the field of dental-based chronological age estimation (from now on, dental age), largely due to its capacity to automate this otherwise laborious process. This development has led to numerous methods, many of which have been outlined in the systematic review by~\cite{vila2023systematic}. One of the earliest methods was proposed by~\cite{cular2017dental}, who developed a statistical model to characterize the morphology and appearance of the third molar and then translate this information into an age estimation. However, since the advent of deep learning in the 2010s, subsequent methods have increasingly relied on this technique, particularly on the capabilities of Convolutional Neural Networks (CNNs), to achieve remarkable outcomes.
In this regard, \cite{banar2020towards} were the first to use a single-step CNN-based pipeline for dental age estimation. This was a significant accomplishment, as the system functioned fully automatically and independently of any specific tooth, with the neural network automatically focusing on the regions of the image deemed most significant. Building upon this foundation, subsequent studies introduced novel approaches to enhance dental age estimation, leading to a better accuracy and reliability than traditional approaches. This is the case of the method proposed by \cite{blanco2023XAS}, one of the methods on which this work is based, and which will be explained in detail in Section \ref{prop}.

\subsection{Explanation Methods}
Despite the notable success of CNN-based systems in dental age estimation, which have surpassed the performance of all previous methods, their explainability remains significantly inferior to traditional approaches such as the one proposed by~\cite{cameriere2006age}. The opaque nature of deep learning architectures, typically comprising a substantial number of convolutional layers, impedes the interpretation of the learned knowledge and, consequently, the reliability of the prediction pipeline. As a result, practitioners utilizing CNNs experience a lack of confidence in the model's decision-making process, leading to predictions~\citep{JMAI8191}. In the absence of knowledge of the rationale behind a decision, there is little confidence that CNNs will make accurate decisions, so their use in high-risk or safety-critical areas is hardly accepted without first developing methods to explain their choices.

The discrepancy between the opacity of automated decisions and their explainability can be mitigated by employing interpretable-by-design transparent models~\citep {rudin2019}. 
Even though the importance of explainability was raised from the beginning of expert systems \cite{mycin}, the earliest explanation generation methods were developed much later ~\citep{Swartout1987, bench1991} for addressing the challenge of explaining the output of expert systems and logic programs. In the context of current XAI, contrary-to-factual (or just counterfactual) explanations are of crucial importance to go beyond summarizing available information about the reasoning behind the system’s output and offer an insight into how alternative outcomes could be reached and appreciated by humans~\citep{stepin2022}. It is also noteworthy that few authors have addressed the challenge of generating explanations in natural language~\citep{cambria2023}. In summary, to pave the way from explainable to trustworthy intelligent systems, researchers must address three levels of explainability~\citep{ali2023explainable}: 
\begin{itemize}
    \item Data explainability: exploratory data analysis methods pay attention to data distribution (e.g., nutrition labels or datasheets for datasets) and can reveal outliers, bias, etc.

    \item Model explainability: according to the level of transparency, models are categorized as ``white-box'' transparent models (e.g., decision trees or rule-based systems), ``gray-box'' models (e.g., Bayesian networks~\cite{BUTZ2022102438} or fuzzy rule-based systems), and ``black-box'' opaque models (e.g., deep neural networks). In the case of white and gray models, it is possible to detect abnormal behavior (e.g., bias) by looking at the model in depth.

    \item Post-hoc explainability: elucidating the reason behind the model's prediction with feature attribution methods (e.g., LIME was implemented by~\cite{lime2016} and it is recognized as one of the pioneer XAI algorithms), visualization methods (e.g., heat maps, class activation maps, CAM, which were proposed by~\cite{gradcam2016} or Gradient-weighted CAM, Grad-CAM, which was implemented~\cite{gradcam2017}), example-based explanation methods, game theory methods (e.g., SHAP was implemented by~\cite{shap2017} and has become one of the most popular XAI methods), etc. In the case of opaque models, explaining predictions is the most common way of finding insights about how a model works~\citep{adadi2018,guidotti_survey_2018}. As an alternative, some authors (e.g.,~\cite{hinton2014}) advocate for model distillation (i.e., extracting knowledge from the opaque model) or use transparent models as surrogates of opaque models (e.g., TREPAN was implemented by~\cite{trepan1995}).
\end{itemize}

In the context of deep learning neural networks, most explainers employ a combination of distillation and visualization methods~\citep{ras2022}. In the specific context of CNNs, which are engineered to address image-related problems, numerous methods exist for generating explanations and enhancing the transparency of CNNs~\citep{haar2023analysis}. Explanations can be elucidated by deconvolution~\citep{zeiler2011} or deep Taylor decomposition~\citep{montavon2017}. Additionally, decision trees can be distilled from CNNs~\citep{zhang2019}. Nevertheless, the most popular methods in practice are visual explanations of CNNs, such as heat maps and saliency maps generated with Grad-CAM or SHAP.

\section{AgeX: an NLG System for the Automatic Generation of Textual reports about dental age estimation}
\label{prop}
This section presents the AgeX intelligent system, which automatically generates descriptive narratives to explain dental-based chronological age estimation. We have followed an interpretability/trustworthiness by design approach for emphasizing this particular aspect. Among the existing options for elucidating a given opaque model, natural language textual reports are particularly advantageous in a clinical context, as they serve as a human-friendly medium for conveying findings derived from data to clinicians. To strike a balance between naturalness and faithfulness, we discarded Large Language Models (LLMs), as their hallucination propensity and inherent reliability, consistency, and interpretability limitations~\citep{NEJMp2405999, Ullah2024} render them unsuitable for dental experts. Furthermore, LLMs are trained on extensive datasets, leading to responses based on correlations and patterns discovered in the data rather than on an understanding of domain expertise. Consequently, deploying LLMs in high-stakes and critical domains such as healthcare, where accuracy is paramount, poses significant risks due to their susceptibility to producing inaccurate or misleading medical advice. 

\subsection{Dental-based chronological age estimation methods}
\begin{figure}[!hbt]
    \centering
    \includegraphics[width=0.75\textwidth]{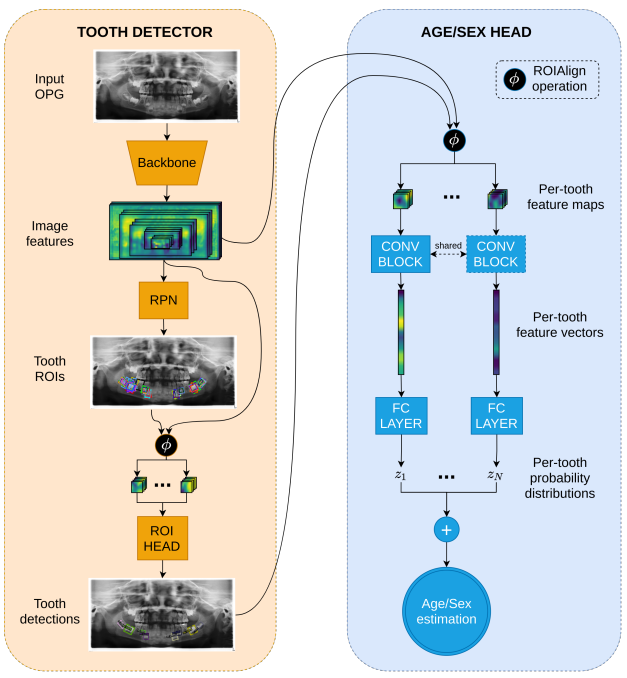}
    \caption{Two-stage CNN-based method for dental age estimation~\citep{blanco2023XAS}. The first convolutional network detects teeth in the input OPG. This information is used along with detected image features as input for the second network, which estimates the patient’s sex and age}
    \label{fig:pipeline-vila-blanco}
\end{figure}

The AgeX system requires patients' dental and chronological age to generate narratives. An external system generates this data. This system, a Opaque Dental Age Estimation System, involves a two-stage CNN approach for chronological age estimation, as illustrated in Figure~\ref{fig:pipeline-vila-blanco}.

In the first stage, the network is configured to detect the specified mandibular teeth (specifically the molars, premolars, and canines). This is achieved by implementing a Faster-RCNN-like architecture that extracts the most salient image features and identifies the tooth-bounding boxes. Note that these boxes are oriented (and therefore not aligned with the X-Y axes), which optimizes the ratio of tooth-background pixels within each box. 
In the second stage, the previously extracted tooth patches are processed with a second CNN to estimate the per-tooth chronological age. However, the output of this stage is not a single number but rather a parameterized probability distribution $z_t, t \in T$, where $T$ is the set of detected teeth. In this context, the network provides a normal distribution $z_t \sim \mathcal{N}(\mu_t,\,\sigma_t^{2})$ which yields two values: a point estimate ($\mu$) and a measure of uncertainty ($\sigma$). The larger the $\sigma$, the greater the uncertainty of the prediction. 
The per-tooth predictions are then aggregated with their associated uncertainty using the following formula:

\begin{equation}
    \label{eq:aggregation}
    \hat{y} = \sum_{t \in T} \mu_t \cdot w_t \qquad,\qquad w_t = \frac{1/[(\sigma_t^2)^p]}{\sum_{t \in T} 1/[(\sigma_t^2)^p]}
\end{equation}
where $p$ is a penalty parameter used to increase the effect of uncertainty in the aggregation.

The present study has quantified the uncertainty (measured by the standard deviation of the data distribution) inherent in the age estimation performed for each tooth of a patient. The overall estimation uncertainty is calculated using the quantification model proposed by Zadeh for relative quantifiers:
\begin{equation}
Z_{Q_{rel}}(A) = \mu_{Q_{rel}} \left( \sum_{t \in T} \frac{a_t}{n} \right)
\label{eq:zadeh_rel}
\end{equation}
where $n$ is the number of dental pieces available for a patient and $a_i$ is the reliability of each tooth. The reliability is calculated by applying fuzzy sets based on the uncertainty associated with the age estimate for each tooth ($\sigma_t$ in \eqref{eq:aggregation}).

This explanation of reliability based on the per-tooth uncertainty is accompanied by a discussion of which dental pieces are considered to be the most predictive, i.e., those with the lowest estimated standard deviations. Determining the most predictive teeth for each patient is achieved by implementing the density-based spatial clustering of applications with noise (DBSCAN) method proposed by~\cite{DBSCAN1996}. Specifically, clustering is applied to the $\sigma_t$ values, specified as a minimum of one element per category, with a separation parameter empirically set at $0.03$. DBSCAN then organizes the data into clusters based on the similarity of the data values. Consequently, the teeth with the most reliable measurements fall within the first cluster, exhibiting the most minor standard deviations. 

Subsequently, the Grad-CAM algorithm generates a visual explanation of the age estimation, highlighting the regions of the dentition that the CNN focused on. This enables dental professionals to evaluate which tooth is more relevant to a particular prediction and which regions within each tooth contributed the most.

\begin{figure}[!ht]
    \centering
    \includegraphics[width=0.55\textwidth]{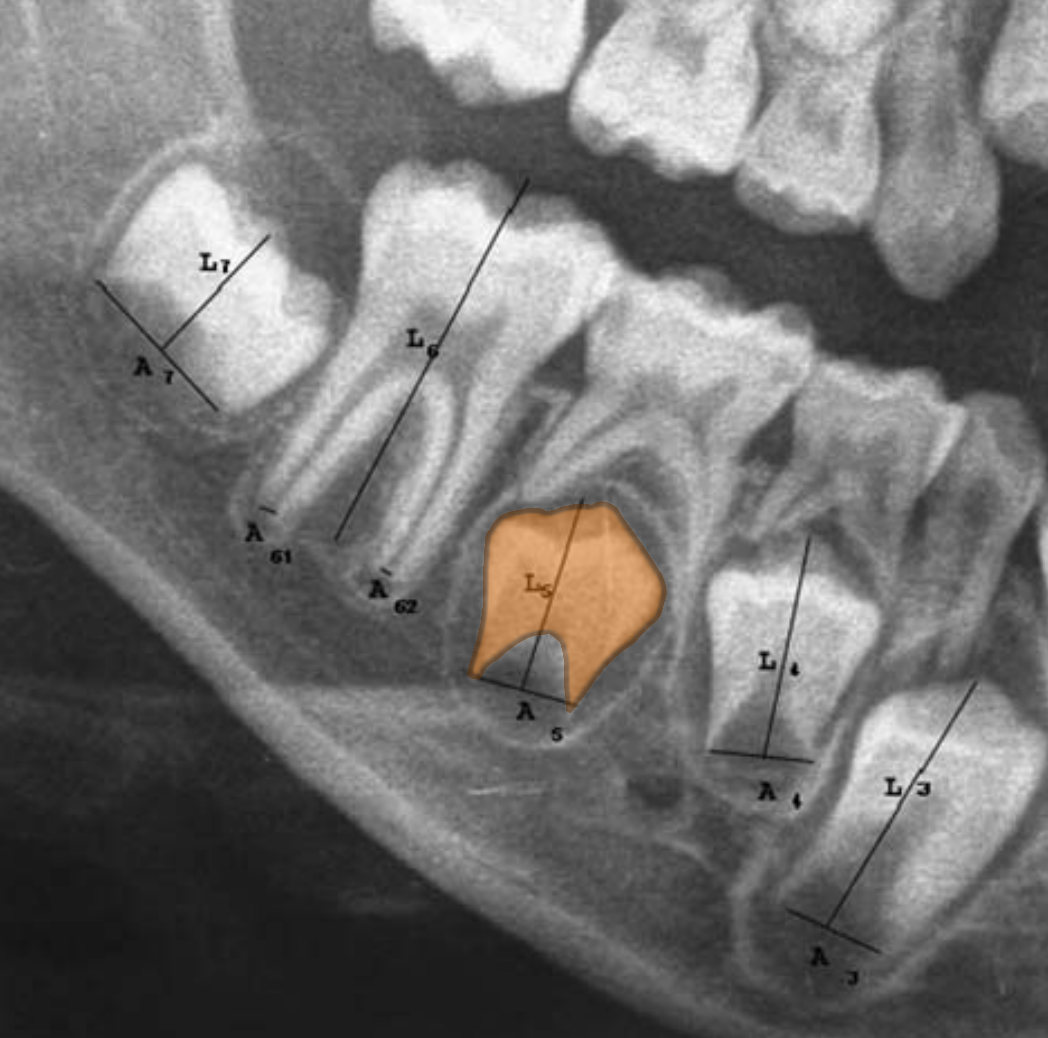}
    \caption{Measurements used to compute $CSM(4i)=A_i/L_i$, $i \in \{1,...,7\}$ by~\cite{cameriere2006age}. $A_i$ correspond to the apex openness, while $L_i$ correspond to each tooth height. The image shows the mandibular right quadrant of a patient's dentition, with tooth 45 (second premolar) highlighted}
    \label{fig:cameriere_measurements}
\end{figure}

To provide a more comprehensive visual explanation (dental images), dental experts recommended the development of a textual explanation (narratives) that is as straightforward and intuitive as possible, similar to that associated with the well-known transparent estimation method proposed by~\cite{cameriere2006age}. This method is based on tooth measurements in the fourth quadrant of the dentition, as illustrated in Figure \ref{fig:cameriere_measurements}. It is easy to explain reasonably and understandably both the application of the method and the influence of the patient's dental characteristics on the estimation of age, which is done by the following linear regression model:
\begin{equation}
\label{eq:age_cameriere}
\begin{aligned}
Age &= 8.971 + 0.357 \cdot g + 1.631 \cdot CSM(45) \\ 
& + 0.674 \cdot N_0^4 - 1.034 \cdot s - 0.176 \cdot s \cdot N_0
\end{aligned}
\end{equation}
\begin{equation}
s = \sum_{i} {CSM}(4i)\quad i=1,...,7
\label{eq:s_cameriere}
\end{equation}
where $g$ is the biological sex of the patient (scored $1$ for boys and $0$ for girls), $CSM(45)$ is the Cameriere’s Standardized Measurement (i.e., the openness of the apex divided by the tooth height) of tooth $45$ (the second premolar), $N_0$ is the number of teeth whose apices are entirely closed, and $s$ is the sum of the $CSM$ in each tooth, as described in Equation \ref{eq:s_cameriere}. The coefficients and the constant must be adjusted to the population, since different populations show different developments~\citep{tompkins1996human}. 

The manual method proposes a less precise but explainable methodology as it is a linear regression. This allows dental professionals to understand and trust the estimations provided. On the contrary, although the estimates made by the automatic method are generally much more accurate than those made by the manual method, we can really only reason about the error it makes in predicting the age of patients.

\subsection{Textual Explanations Construction}
AgeX implements an NLG trustworthiness module to generate narratives. It follows the generation pipeline defined by \cite{reiter2000}, which comprises three stages: 
\begin{figure}[!ht]
    \centering
    \includegraphics[width=0.85\textwidth]{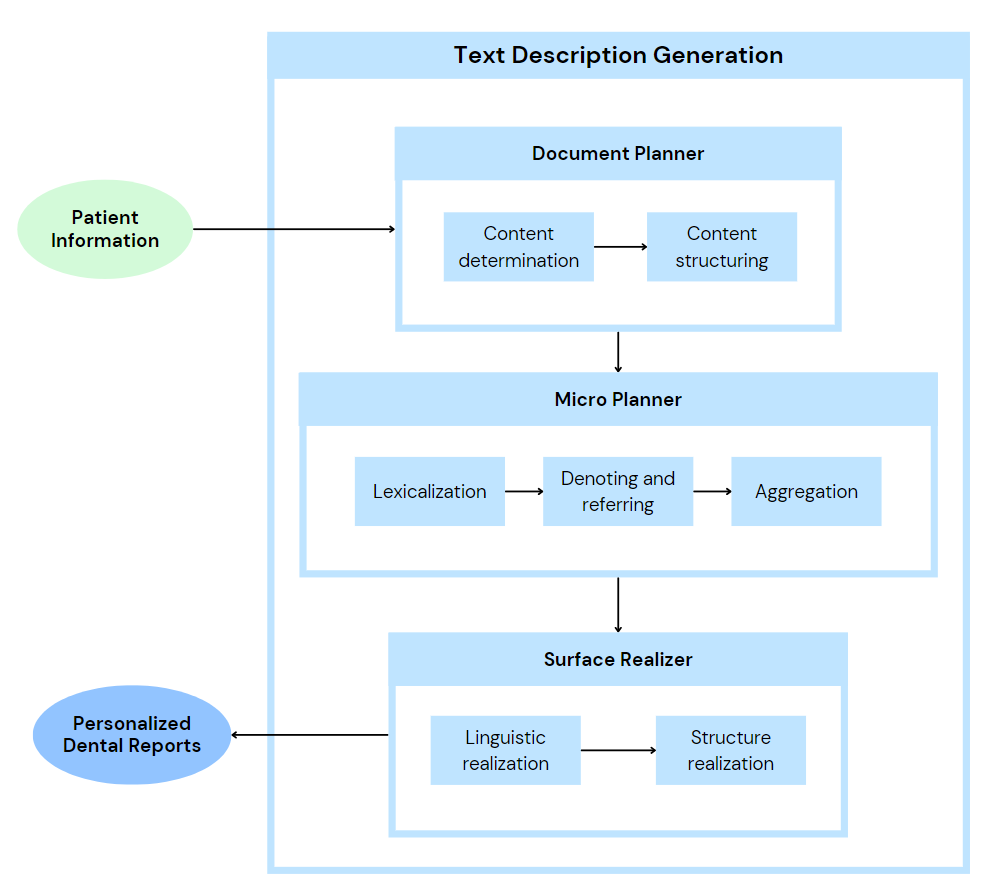}
    \caption{The Text Description Generation system in the core of AgeX. The system uses patient information to generate narratives. This information is used to determine what to say and how to structure the content. Then, the system designs how to convey the information and finally generates the correct text, resulting in personalized dental reports}
    \label{fig:pipeline}
\end{figure}
\begin{enumerate} 
    \item Document Planning. This stage involves the strategic generation of narratives regarding content determination and structuring. We need to specify what to say in the writing. First, we asked a dental expert to write some sample reports manually. Then, from the given examples, we created a set of generation rules and a dynamic report template that were validated and curated by the expert.  
    
    \item Micro Planning. At this stage, we must determine how to convey the information (i.e., how to say things in writing). Following the guidelines established by our dental expert, we defined the content and structure of two types of sentences:
    
    \begin{itemize}
        \item Static sentences are those that do not change between patient cases because they contain general information about the chronological age estimation process or the explanation method.
        \item Dynamic sentences are expected to change for each patient, since they rely on data retrieved from the database. The designed rule-based system determines which sentences are generated in relation to the available data for each particular case (e.g., using female child or male child following the patient's sex if it is known, or showing/hiding some particular relevant details under the patient's clinical record).
    \end{itemize}

    The rule-based system generates a meaningful report by applying rules with conditional statements. In addition, fuzzy quantifiers translate quantitative data into descriptive, qualitative terms~\citep{ZADEH1983149}. Fuzzy quantifiers are linguistic terms (e.g., none, a few, some, most, all) that provide approximate, human-friendly descriptions of numerical values. Using these quantifiers enables the generation of more natural and readable narratives. For example, instead of specifying an exact number that may be overly technical or difficult to decipher, we use fuzzy quantifiers to categorize the developmental status of a patient’s teeth. The integration of static and dynamic sentences logically and coherently results in formulating the patient’s dental report.
    
    \item Surface Realization. The final stage is to apply syntactic and morphological rules to generate correct text. We have to say things. The SimpleNLG~\citep{simplenlg} and SimpleNLG-ES~\citep{aramossoto2017adapting} support linguistic realization.
\end{enumerate}

\begin{figure}
    \centering
    \includegraphics[width=0.95\linewidth]{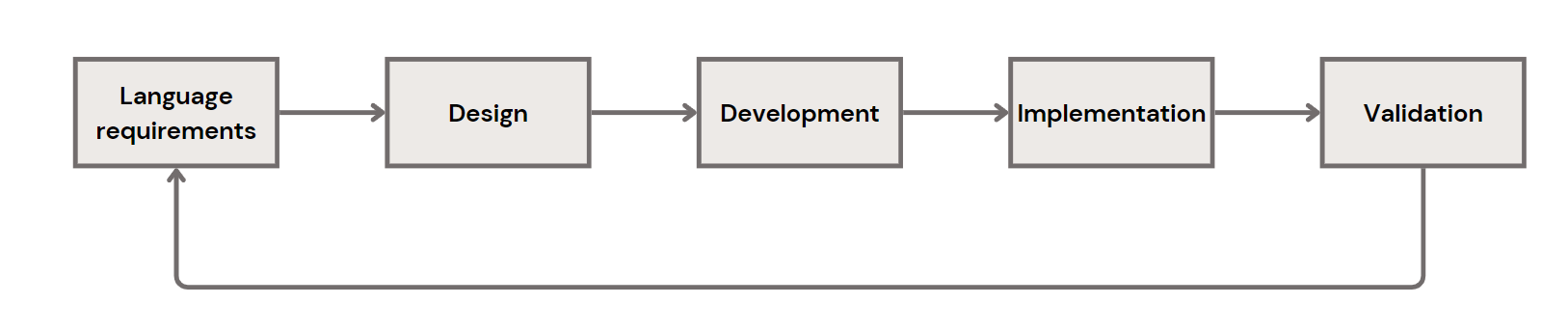}
    \caption{Diagram illustrating the iterative and incremental development process of AgeX, with continuous feedback and validation at each stage}
    \label{fig:sw}
\end{figure}

The experts provided the language requirements for our system to determine the content of the narratives and the information that should be included. We followed an iterative and incremental development process, where, as part of this methodology for creating the narratives, the validation of the explanations with the dental expert is always included (Figure~\ref{fig:sw}).

As previously stated, the requirements for this application were established by experts, with a 100\% demand for the accuracy of the texts when describing each patient's dental condition. In this context, the design of a rule-based Data-to-Text (D2T) system is the most suitable option, as the issues previously identified for neural End-To-End NLG systems do not occur when applying D2T approaches. This is due to the fact that the precision and accuracy of the generated descriptions are ensured by design. Furthermore, when designing rule-based D2T systems, multiple iterations and tests are conducted to guarantee the quality and appropriateness of the generated texts. According to the prevailing best practices~\citep{howcroft-etal-2020-twenty, Lee2021}, the most arduous form of testing for these systems is final validation by human experts. This standard validation methodology cannot be supplanted by automatic metric-based assessments or comparisons with other generators.
 
The narratives generated in AgeX can be one of three types:
\begin{itemize}
    \item Introduction. Statement common to both descriptions (short and detailed). This sentence specifies the biological sex of the patient (``a [male\slash female]''), their real chronological age and the type of dentition (``dentition [deciduous\slash mixed\slash permanent]'') they should have according to their age (``a [child\slash adolescent\slash adult]'').

    \item Short description. The dental age estimate is given with an explanation of the result. The explanation is based on the number of fully developed teeth (``[none\slash few\slash most\slash all]'') in the right mandibular quadrant and the opening of the apex of the right mandibular second premolar (``[fairly\slash completely] closed'' or ``[fairly\slash quite] open'').

    \item Detailed description. In this case, the sentences are organized into three paragraphs.
    \begin{enumerate}
        \item The dental age estimate is given with an explanation of the result. The explanation is based on the number of fully developed teeth (``[none\slash few\slash most\slash all]'') in the right mandibular quadrant (how many out of 5) and the opening of the apex of the right mandibular second premolar (``[fairly\slash completely] closed" or ``[fairly\slash quite] open''). It is stated that the age estimate was calculated using the opaque method by \cite{blanco2023XAS}. 
        \item Analysis of the difference between real and estimated chronological age. 
        \begin{itemize}
            \item The real age of the patient is mentioned again.
            \item The difference between the real and estimated age is displayed, calculated as shown in Equation~\ref{eq:diff_age}.
            \begin{equation}
                error = \abs{ \frac{age_{real} - age_{opaque}}{age_{real}}} \cdot 100
                \label{eq:diff_age}
            \end{equation}
            If this error is significant (greater than 1\%), then we use quantifiers ``[under\slash over] estimated age exists''. If it is lower, then we generate ``the estimate is accurate''.
            \item The reliability of the estimate is described (``the reliability of the estimate is [low\slash high]''), indicating which teeth were most predictive for its calculation. 
            If the reliability of the estimate is normal, it is not mentioned and the information about the most predictive teeth is provided directly.
        \end{itemize}

        \item The calculation of dental age using the opaque method is described. It explains how dental age is estimated, how heat maps are obtained, and how the color assigned to each tooth is determined. This paragraph is static and contains the following sentences: 
            \begin{itemize}
            \item ``The patient's age was estimated using the Vila-Blanco method, an automated approach based on deep learning.''.
            \item ``This method detects the patient's teeth in their OPG and calculates the probabilistic distributions associated with the chronological age of each tooth.''.
            \item ``The patient's age is estimated by aggregating the dental age estimates for each of the patient's teeth.''.
            \item ``With the information inferred by the automatic method, the heat-map of the corresponding OPG is shown in the figure.''.
            \item ``For this purpose, each pixel of each mandibular tooth is assigned a color according to its importance for the chronological age estimation.''.
            \end{itemize}
    \end{enumerate}
\end{itemize}

The generated report shown to a dental professional is a combination of the introduction with a short or detailed description, depending on the preference of the dental expert using the AgeX system. The generation of narratives requires the acquisition of patient-specific data. Anonymized data is collected from external sources, such as medical records, and stored in the database. Each patient is identified by a unique ID, and their associated record contains specific details about their real chronological age, sex, and dental measurements. This information is supplemented with dental data and heat maps generated by the Vila-Blanco system. 

\begin{figure}[!ht]
    \centering
    \includegraphics[width=0.85\textwidth]{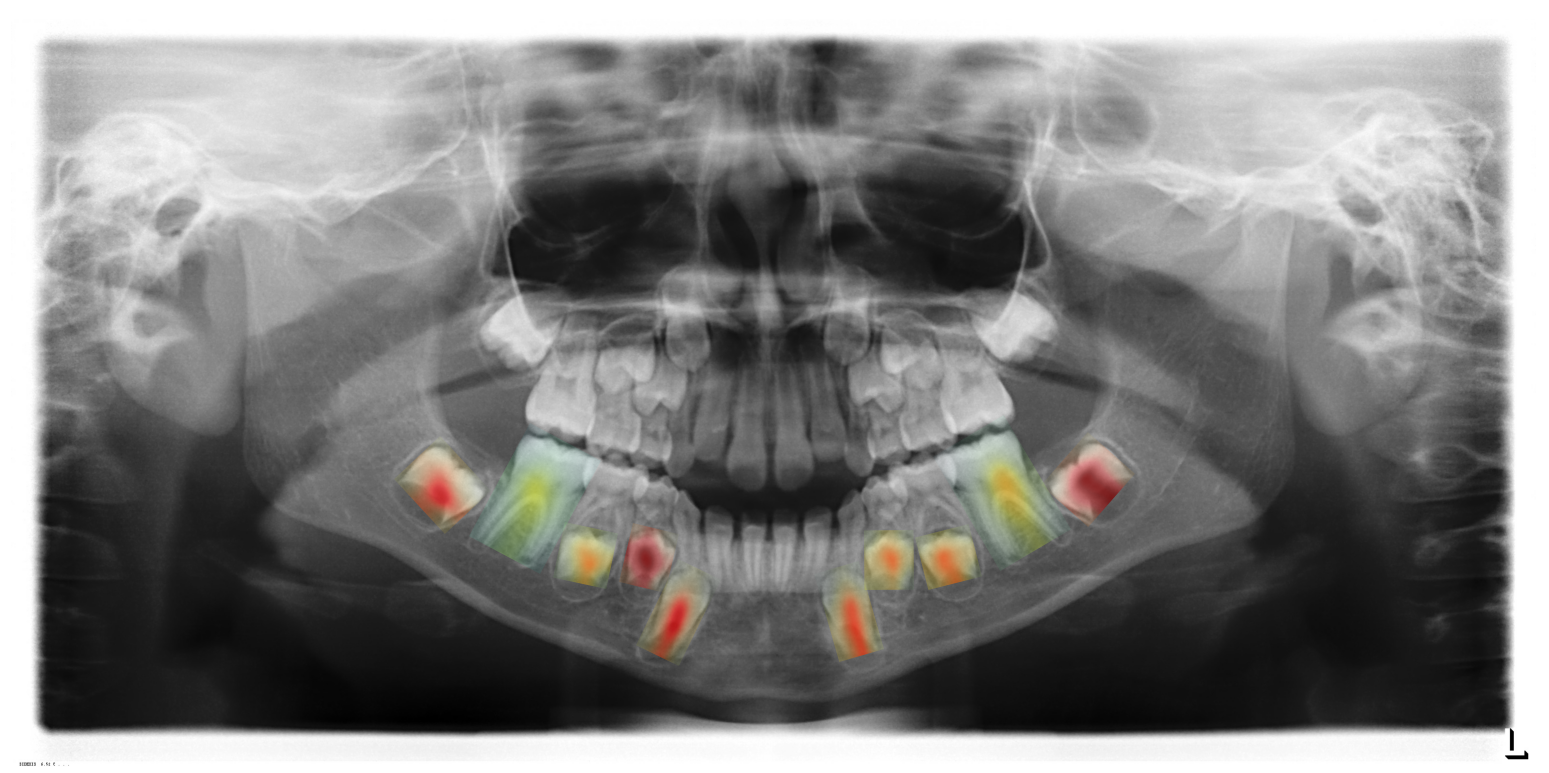}
    \caption{Example of a heat-map created from the OPG of a patient. In the image, the patient's permanent mandibular teeth are highlighted. The most predictive teeth are marked in red, while the least predictive ones are in green. The uncolored lower teeth are primary (baby) teeth}  
    \label{fig:mapacalor2324}
\end{figure}

As an illustrative example, we show in what follows the textual descriptions generated with AgeX for the patient in Figure~\ref{fig:mapacalor2324}.
The introduction is:
\begin{displayquote}
\small\textit{A female child with a real chronological age of 7 years and 2 months should present mixed dentition.}
\end{displayquote}
The short description is: 
\begin{displayquote}
\small\textit{The dental age estimate is 7 years and 9 months because she has no fully formed teeth in the right mandibular quadrant and the apex of the right mandibular second premolar is quite open.}
\end{displayquote}
and the detailed description is:
\begin{displayquote}
\small\textit{The dental age estimate, according to the Vila-Blanco et al. method, is 7 years and 9 months because she has none fully formed teeth (0 of 5) in the right mandibular quadrant and the apex of the right mandibular second premolar (45) is quite open.}
\small\textit{The real chronological age is 7 years and 2 months. Therefore, there is an overestimation (7 months) with respect to the real age. The reliability of this estimate is high, with the most predictive teeth being 44 and 37.}

\small\textit{The patient's age was estimated using the Vila-Blanco method, an automated approach based on deep learning. This method detects the patient's teeth in their OPG and calculates the probabilistic distributions associated with the chronological age of each tooth. The patient's age is estimated by aggregating the dental age estimates for each of the patient's teeth. With the information inferred by the automatic method, the heat-map of the corresponding OPG is shown in the figure. For this purpose, each pixel of each mandibular tooth is assigned a color according to its importance for the chronological age estimation.}
\end{displayquote}

The heat-map of this patient is shown in Figure \ref{fig:mapacalor2324}. Original Spanish texts are available on \ref{app1:original_texts}. 

\subsection{AgeX architecture}
\begin{figure}[!ht]
    \centering
    \includegraphics[width=0.87\textwidth]{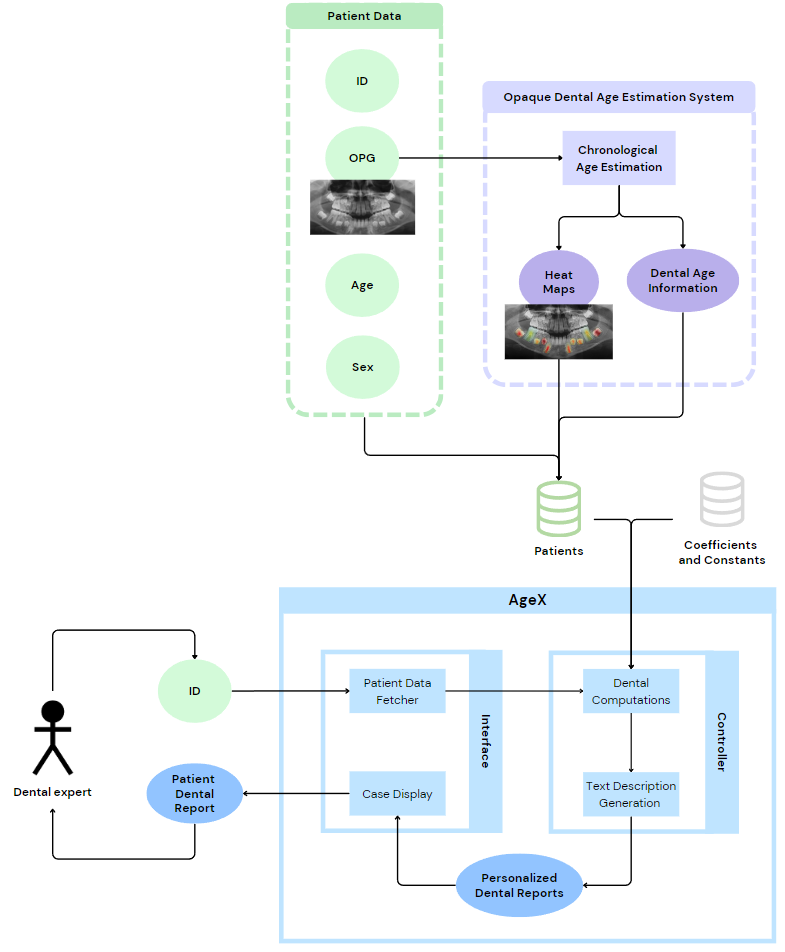}
    \caption{Process of capturing, storing and accessing patient information in AgeX. The dental expert enters the patient ID to retrieve the information she wish to consult. The system searches for the dental data, performs the necessary calculations, and generates the narratives. These descriptions, along with the corresponding OPG and heatmap, are presented to the expert. The data required for the calculations are obtained from the database containing information on all registered patients, as well as from the database containing coefficients and constants for the equations}
    \label{fig:data_flow}
\end{figure}

The architecture of AgeX is illustrated in Figure~\ref{fig:data_flow}. Dental professionals who wish to access patients' data are required to enter their respective ID into the system. Subsequently, the system accesses the Patients database and retrieves the stored information concerning the patient.  
Subsequently, a report is automatically generated by the Controller module and displayed through the Interface module. The report contains multimodal data (i.e., text and images). It provides a synopsis of the salient details relevant to the specified ID, based on data from the Coefficients and Constants database. 

AgeX transforms structured data into trustworthy reports, thereby providing clinicians with accurate and comprehensible dental age estimates for each case under study. The system's core (i.e., the Controller) is responsible for interpreting the patient data. It employs predefined rules (co-designed with a dental expert) to generate justifications that enhance the comprehensibility of the underlying dental age estimation method. Besides this, it also uses a rule-based NLG system that relies on well-defined, transparent algorithms for data analysis and clear guidelines derived from established medical knowledge for data interpretation. This approach enables safe, consistent, and controlled text generation following clinical standards and protocols. 
Designed with these principles in mind, the AgeX system ensures predictability, reliability, and alignment with medical standards, providing a reliable foundation for clinical decision support.

The NLG system within the Controller in Figure~\ref{fig:data_flow} is essential for ensuring the system's reliability. This trustworthiness module is further delineated in Figure~\ref{fig:pipeline}. Figure~\ref{fig:data_flow} showcases the Opaque Dental Age Estimation System~\citep{blanco2023XAS}.

\section{Experimental Validation} \label{exp}
We conducted a series of experiments to validate our research hypothesis ``an interpretable, transparent method can be used to explain the results generated by an automatic, opaque method, in such a way that the AgeX system is equipped with a trustworthy, automatic, high-performance approach that sufficiently reliable to be used in a realistic dental age estimation scenario''. This section explains how the experiments were designed.

\subsection{Dataset Description}\label{sec:dataset}
\begin{figure}
    \centering
    \includegraphics[width=0.7\linewidth]{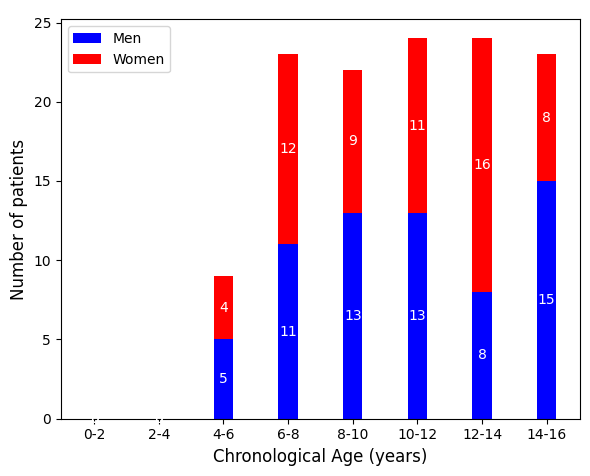}
    \caption{Age distribution of patients by sex, representing the data from the dataset described in Section~\ref{sec:dataset}. We do not have data for patients under the age of 4. For other age groups, the number of dental cases for men and women is fairly balanced, except between ages 12-14, where there are twice as many female cases, and between ages 14-16, where there are nearly twice as many male cases. For each dental group except for 4-6, we have at least 20 dental cases}
    \label{fig:ages}
\end{figure}

The dataset utilized for the validation of the proposed approach consists of 125 dental panoramic images provided by the Oral Sciences Research Group (OSRG) at the University of Santiago de Compostela, with the patient's age and sex known. These images were collected in accordance with the ethical standards of the University of Santiago de Compostela Research Ethics Committee and the 1964 Declaration of Helsinki and its subsequent amendments~\citep{helsinki2013}. The following details are provided for each of the age estimation methods considered in this study. On the one hand, the apex opening measurements and the height of each tooth were manually annotated to apply the method proposed by~\cite{cameriere2006age}. On the other hand, the dental age prediction and the heat-map showing the most relevant areas in each image for making such prediction were obtained by means of the method proposed by~\cite{blanco2023XAS}.
In order to achieve the objectives of homogenization and simplification of the reference to specific teeth, the nomenclature of the Federation Dentaire International (FDI) is employed. This nomenclature assigns a unique numerical label to each tooth~\citep{garg2010textbook}.
Figure~\ref{fig:ages} provides a detailed description of the age and sex distribution of the dataset. Specifically, it comprises a set of 125 patients (60 females and 65 males) aged between 5 and 16, who are all of Spanish-Caucasian descent. The dataset was restricted to this age range, since it covers the period in which most of the teeth are still developing, and therefore, the chronological age estimation is more accurate~\citep{prieto2005evaluation}.

For each patient, the following information is available:

\begin{itemize}
    \item OPG image. The panoramic radiograph in a readable image format.
    \item Patient information.
    \begin{itemize}
        \item age. Chronological age of the patient, expressed in days.
        \item sex. Biological sex of the patient (male/female).
    \end{itemize}

   \item Per-tooth manual measurements. Annotation of relevant distances performed individually for each tooth by experienced dental experts. We refer to each tooth as $xi$, where $xi$ is the number of the permanent mandibular tooth ($x$=quadrant, $i$=position) according to the FDI nomenclature. Note that only the posterior mandibular teeth were annotated (44 to 48 in the right quadrant and 34 to 38 in the left quadrant), following the recommendations in the literature \citep{vila2023systematic}.
    \begin{itemize}
        \item $\bm{xi}$\_height. Height of each dental piece, expressed in pixels. 
        \item $\bm{xi}$\_width. Width of each dental piece, expressed in pixels.
        \item $\bm{xi}$\_ratio. Ratio between $xi$\_height and $xi$\_width.
        \item $\bm{xi}$\_came1. The aperture of the tooth apices of each dental piece, expressed in pixels. A value of $0$ indicates that the first tooth root (the only one in the case of unirradicular teeth) is fully developed.
        \item $\bm{xi}$\_came2. The second opening of the tooth apices is expressed in pixels. Used only for molars as they are multi-radicular. A value of $0$ indicates that the second tooth root is fully developed.
        \item $\bm{xi_{CSM}}$. Cameriere's Standardized Measurement of the dental piece, calculated as described in Section \ref{prop}.
    \end{itemize}

    \item Per-tooth automatic predictions. Predictions are based on the information from each tooth that is extracted by the Vila-Blanco method. We refer to each tooth as $t \in T$, being $T$ the set of detected teeth. In accordance with the manual measurements, this method is trained only to detect the posterior mandibular teeth.
    \begin{itemize}
        \item $\bm{\mu_t}$. Mean of the estimated distribution of dental age. It can be seen as a point estimation of age.
        \item $\bm{\sigma_t}$. Standard deviation of the estimated distribution of dental age. It can be seen as a measure of the uncertainty of the estimate.
    \end{itemize}

    Note that for a particular OPG, the set of teeth with manual measurement may not be the same as the set of teeth with automatic predictions, since Vila-Blanco's method relies on a tooth detector that may miss some existing teeth. AgeX uses exclusively the set of teeth with automatic predictions. 

    \item Heat-map. An image where a color map is superimposed upon the original OPG. The color information is indicative of the importance of each pixel belonging to a detected tooth (as determined by the Vila-Blanco method) in the chronological age estimation process. Consequently, pixels that are the most indicative of a tooth appear in a deep red hue, while pixels that are the least indicative appear in a green hue.
\end{itemize}

Since all the images in this dataset have identical pixel dimensions, measurements were conducted in pixel units for consistency.

\subsection{Statistical Validation}
\label{sec:statistical_validation}

To prove the suitability of AgeX's dental age estimation method, two statistical analyses were conducted.

First, a correlation study was performed to ascertain the relationship between the dental attributes utilized by the transparent Cameriere's method and the information generated by the opaque Vila-Blanco's method. Specifically, the Pearson correlation coefficient was used to assess the linear dependence between each pair of quantitative random variables. The objective of this correlation analysis is to determine whether the information obtained through the opaque Vila-Blanco method can be considered equivalent to that provided by the transparent Cameriere method, so we can use supervised learning to facilitate the development of a regressor that estimates the dental age based on the opaque Vila-Blanco method and employs the transparent Cameriere method as a surrogate explainer. 
To this end, two regression models were evaluated: a linear regression model and a Multilayer Perceptron (MLP). Both regressors were trained to estimate the Cameriere's Standardized Measurement (CSM) based on biological sex, expert annotated dental ratio ($xi\_ratio$ in Section \ref{sec:dataset}) and the chronological age estimated by Vila-Blanco ($\mu_t$ in \ref{sec:dataset}). This procedure is initially performed for tooth $45$, as it is the one utilized by Cameriere. Then, this procedure is repeated for the remaining dental pieces. 

Second, a reliability study was performed for the estimations of the opaque method, aiming to provide an approximation of the uncertainty associated with the estimation to dental experts. To this end, the distribution of uncertainties in the per-tooth age estimations was characterized through the mean and standard deviation. The results of this analysis are noted for future reference.

\subsection{Human manual validation by dental professional experts}
\label{sec:questionnaire_based_validation}
An intrinsic validation of the AgeX system was conducted following the evaluation guidelines for NLG systems (\cite{howcroft-etal-2020-twenty}). 
In accordance with these guidelines, an anonymous group of seven dental experts validated a selection of narratives generated by AgeX. These experts answered the questions given in a questionnaire that the authors meticulously designed. 

The dental expert who is a co-author of this paper contributed to the design of the validation questionnaire and selected four medical validation cases on the basis of their diversity and interest\footnote{Following the best practices in validation, this expert did not participate in the evaluation.}. 

For the human evaluation, we selected four patient cases and developed three distinct scenarios for each case. The rationale behind creating three scenarios was to assess how experts responded to the amount of information provided in each case. Specifically, we aimed to determine whether the introduction of additional information was beneficial and well-received, or if it was deemed unnecessary. In the first scenario (S1), dental experts were provided solely with the specification of the dental age estimation on the patient's OPG, devoid of any supplementary explanation. In the second scenario (S2), dental experts were provided with a concise explanation of the dental age estimation of the patient's OPG. The third scenario (S3) entailed the provision of a detailed explanation of the dental age estimation on the heat-map obtained from the analysis of the patient's OPG.

The panel of dental experts was tasked with providing honest answers to five questions, using a five-point Likert scale to rate each statement intended to assess their level of satisfaction with each scenario, and to select the scenario they considered most suitable for regular use in a clinical setting based on their experience. To be specific, with the statements of Likert scale we evaluate the following aspects of the automated generated narratives:
\begin{enumerate}
    \item Adequacy. The textual description uses appropriate terminology to describe the patient's dentition. 
    \item Clarity. The description is easy to understand. 
    \item Informativeness. The textual description provided contains sufficient information.  
    \item Usefulness. The textual description provided includes useful information. 
    \item Consistency. The textual description is consistent concerning the image accompanying it. 
\end{enumerate}

Finally, we asked which of the three scenarios seems most realistic or feasible for regular use in a clinical setting, based on their own experience. 

The data collected were processed and analyzed in compliance with the European General Data Protection Regulation (GDPR) provisions.
In addition, to identify potential risks and issues for improvement in alignment with the European principles for trustworthy AI, a self-assessment of the trustworthiness of AgeX was conducted using the Assessment List for Trustworthy Artificial Intelligence (ALTAI) proposed by the~\cite{altai_website}. The ALTAI questionnaire is a dynamic instrument with questions that are dependent on the answers provided in previous questions. The objective of the questionnaire is to evaluate whether an AI system being developed, deployed, procured, or utilized meets the following seven principles outlined in the European Ethics Guidelines for Trustworthy AI~\citep{ethics_guidelines_2019}, which laid the groundwork for the AI Act~\citep{EURegulation1689_2024} by providing ethical guidelines and policy recommendations that shaped the regulatory framework of the AI Act:
\begin{itemize}
    \item Human Agency and Oversight discusses the importance of AI systems in supporting fundamental rights, human autonomy, and decision-making, emphasizing the need to respect human autonomy.

    \item Technical Robustness and Safety emphasizes that stability, reproducibility, and resilience to attacks are essential to achieve trustworthiness.

    \item Privacy and Data Governance highlights the importance of taking care of access to data and integrity of data when designing and deploying AI systems.

    \item Transparency pays attention to traceability, explainability, and open communication regarding the AI system's limitations.

    \item Diversity, Non-Discrimination and Fairness emphasize the need for inclusion and diversity throughout the AI system’s lifecycle. 

    \item Societal and Environmental Well-being highlights the need to consider broader society, sentient beings, and the environment as stakeholders throughout the AI system's lifecycle.

    \item Accountability refers to mechanisms to ensure the auditability of AI systems in the search for liability in their development, deployment, and use.
\end{itemize}

ALTAI provides guidance on comprehending the concept of Trustworthy AI, identifying potential risks associated with AI systems, and assessing the impact of these systems on society, the environment, and vulnerable groups. Specifically, it assists organizations and individual developers in evaluating the trustworthiness of their AI systems. ALTAI consists of filling out a self-assessment questionnaire about each of the seven principles mentioned above. 

\section{Analysis of Results} 
\label{cap:test}
\subsection{Statistical Validation}
\label{sec:stat1}


First, we present the results of the analyses described in Section \ref{sec:statistical_validation}, with a view to demonstrate that both the opaque (automatic) and transparent (manual) methods yield analogous information.




The results of the Pearson's correlation analysis are summarized in Table~\ref{tab:comp}. All coefficients were negative and greater than 0.5 in absolute terms, the great majority over 0.7. First, it is proved that the higher the ratio the lower the estimated chronological age. This is in line with the process of tooth development, which starts with the tooth crown and continues with the development of the tooth root, so it becomes taller over time. In addition, the correlation values between the ratio and the estimation uncertainty indicate that the most substantial modifications in dental anatomy are observed during the initial developmental stages, so the uncertainty of predictions escalates as the ratio diminishes. The same can be applied to CSM-related correlation values, with a key difference. When the apices are fully closed, this measure reaches $0$ and is no longer informative.

Notably, teeth 37 and 47 (mandibular second molars) exhibited the highest correlation values. Although these dental pieces demonstrate the strongest correlation, following the recommendation of our expert, we opted to utilize 45 (right mandibular second premolar), as it shows a robust correlation and is considered a superior predictor of age in the existing scientific literature~\citep{cameriere2006age}. 

\begin{table}[!ht]
    \centering
    \caption{Pearson correlation coefficients computed for each mandibular tooth on the right side (teeth 48 to 44) and left side (teeth 34 to 38) based on their measurements (ratio and CSM) against the estimated age for each tooth (mean and standard deviation -STD-). Each tooth is accompanied by age markers, specifically, the age at which mineralization begins (Miner.) and the age at which root calcification completes (Root Dev.). The highest correlation values for each comparison are highlighted in bold. The tooth suggested that was identified by dental experts as the most predictive is indicated in gray.
    }
        \begin{tabular}{ccccccc}
        \hline
        \multirow{3}{*}{\textbf{\makecell{Dental \\ Piece}}} 
        & \multicolumn{2}{c}{\textbf{Age Markers}} 
        & \multicolumn{4}{c}{\textbf{Pearson correlation coefficient}}  \\ 
        
        \cmidrule{2-7} 
        & \textbf{Miner.} 
        & \textbf{Root Dev.} 
        & \makecell{$\bm{xi}$\_ratio \\ vs $\bm{\mu_t}$} 
        & \makecell{$\bm{xi}$\_ratio \\ vs $\bm{\sigma_t}$} 
        & \makecell{$\bm{xi_{CSM}}$ \\ vs $\bm{\mu_t}$} 
        & \makecell{$\bm{xi_{CSM}}$ \\ vs $\bm{\sigma_t}$} 
        \\ \hline
        
        48                          & 8       & 25      & -0.840              & -0.710             & -0.582            & -0.541          \\
        47                          & 2.5     & 15      & \textbf{-0.878}     & -0.739             & \textbf{-0.890}   & -0.763          \\
        46                          & $<$1    & 10      & -0.726              & -0.585             & -0.753            & -0.601          \\
        \rowcolor[HTML]{C0C0C0} 45  & 2       & 14      & -0.825              & -0.748             & -0.835            & -0.746          \\
        44                          & 1.5     & 13      & -0.862              & -0.786             & -0.766            & -0.682          \\ \hline
        34                          & 1.5     & 13      & -0.844              & -0.712             & -0.850            & -0.741          \\
        35                          & 2       & 14      & -0.830              & -0.703             & -0.855            & -0.731          \\
        36                          & $<$1    & 10      & -0.681              & -0.544             & -0.813            & -0.649          \\
        37                          & 2.5     & 15      & -0.853              & \textbf{-0.810}    & -0.860            & \textbf{-0.811} \\
        38                          & 8       & 25      & -0.845              & -0.738             & -0.669  & -0.588          \\ \hline
        \end{tabular}
\label{tab:comp}
\end{table}

\begin{figure}[!ht]
    \centering
    \includegraphics[width=0.95\textwidth]{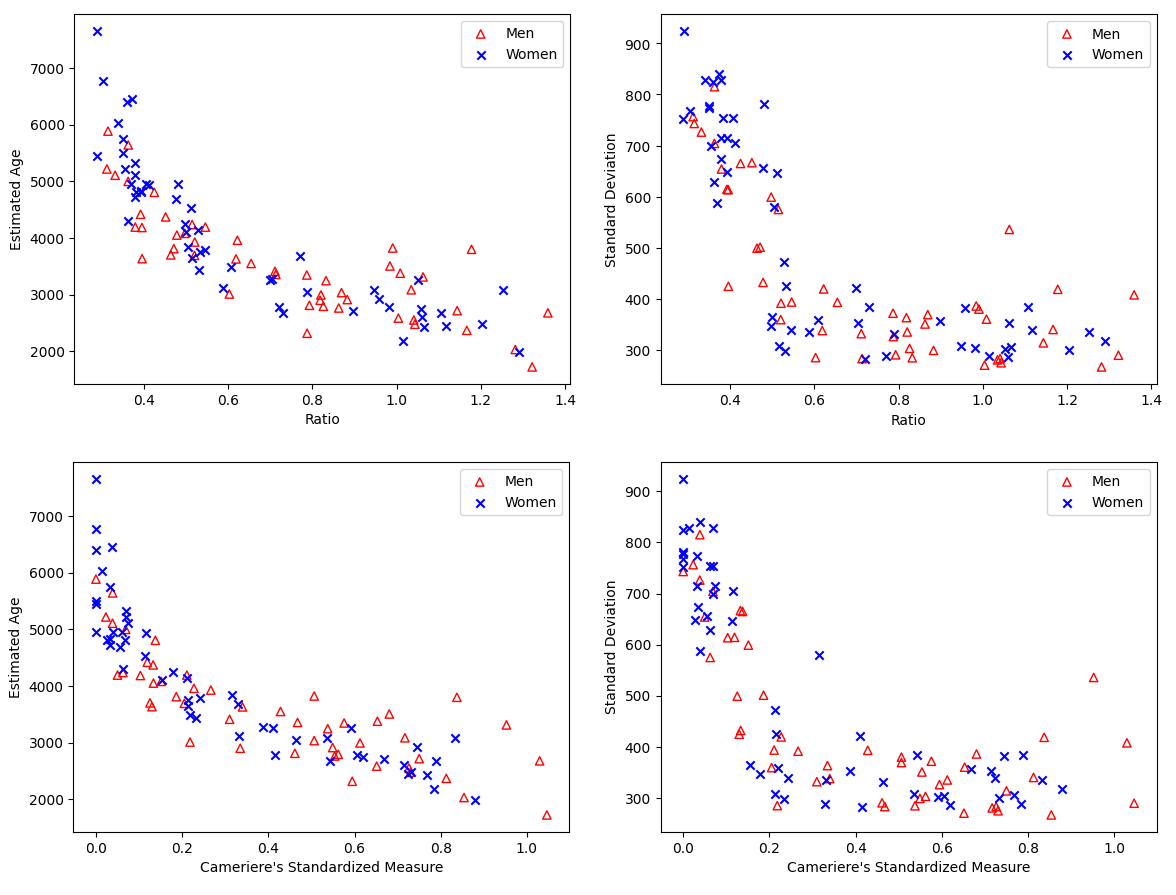}
    \caption{Scatter plots showing the relationship between $\bm{xi}$\_ratio and $\bm{xi_{CSM}}$ with $\bm{\mu_t}$ and $\bm{\sigma_t}$ of dental piece 45 (right mandibular second premolar), grouped by sex. The chronological age estimation is in days. The plots show that the variables are closely related. As $\bm{xi}$\_ratio decreases, both $\bm{\mu_t}$ and $\bm{\sigma_t}$ decrease in a similar manner. However, the decrease is more pronounced for $\bm{\sigma_t}$. The same happens for $\bm{xi_{CSM}}$; when it decreases, both the estimated age and the standard deviation decrease as well. Additionally, the distributions for men and women are similar, indicating consistent patterns between sexes}
    \label{fig:comp_45}
\end{figure}

The relationship between the estimated chronological age and ratio and CSM on this tooth can be further elucidated by examining the scatter plots presented in Fig.~\ref{fig:comp_45}, where no notable differences can be seen between males and females. The data suggest that both ratio and CSM are inversely correlated with age, meaning higher values of these metrics are associated with lower estimated ages. The decreasing standard deviation in both top-right and bottom-right graphs implies more consistent predictions at higher values of the ratio and standardized measure. This decrease in standard deviation as these metrics increase suggests that the metrics become more reliable and consistent predictors of age as the teeth become more formed.

\begin{figure}[!ht]
    \centering
    \includegraphics[width=0.95\textwidth]{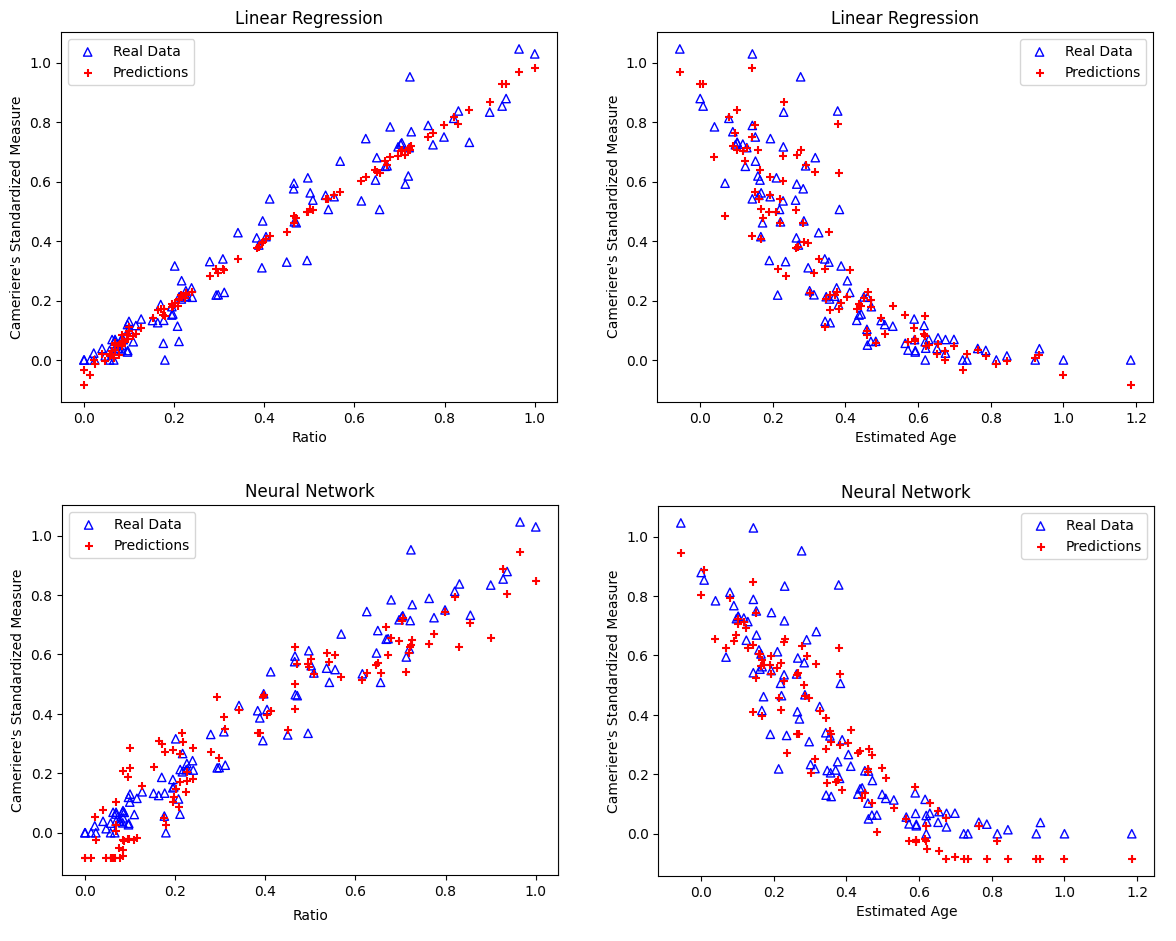}
    \caption{Scatter plots showing the difference between the CSM estimation (by each regressor) and its actual value ($\bm{xi_{CSM}}$). For both methods, the visual representation shows that they can adjust fairly well to the original data distribution}
    \label{fig:models}
\end{figure}

As Cameriere's method utilizes tooth 35 and 45 (second premolars) to estimate chronological age, the regression analysis was conducted with tooth 45 (right mandibular second premolar) as the sole input. 
Figure \ref{fig:models} visually compares the real and estimated values using each regressor, and Table~\ref{tab:res_45} presents a concise overview of the calculated evaluation metrics. The results demonstrate that, while both models perform well, the Linear Regressor consistently outperforms the MLP in all 7 metrics. The Linear Regressor shows superior Explained Variance ($0.946$ vs. $0.933$), Mean Squared Error ($0.005$ vs. $0.007$), and Mean Error ($0.069$ vs. $0.079$). These results suggest that the Linear Regressor provides more accurate and reliable predictions with smaller errors on average. Additionally, the Maximum Error and Average Absolute Error are both lower for the Linear Regressor, indicating fewer extreme errors and better consistency in its predictions. The Coefficient of Determination (R²) is also higher for the Linear Regressor ($0.943$ vs. $0.922$), highlighting its ability to explain a greater proportion of the variance in the target variable. Moreover, the Pearson’s Correlation Coefficient is similar for both models, with the Linear Regressor slightly ahead ($0.973$ vs. $0.971$), reflecting a nearly identical strength of linear relationship between predicted and actual values in both models. While the MLP may be a more complex model capable of capturing non-linear relationships, the results suggest that for this particular dataset, the Linear Regressor is more effective. The Linear Regressor not only delivers superior performance on all metrics but also offers a distinct advantage in terms of transparency and interpretability, which is critical in many practical applications where understanding the model’s behavior is essential.
We conclude that the Linear Regressor is the preferred choice for this dataset. Despite its simplicity, it accurately estimates CSM. Consequently, the lineal regressor will be employed in AgeX. 

\begin{table}[!ht]
    \centering
    \caption{Results of the evaluation of the regressors for the estimation of the CSM of the teeth 45 (right mandibular second premolar).}
        \begin{tabular}{lcc}
        \hline
        \textbf{Metric} & \textbf{Linear Regressor} & \textbf{MLP} \\ \hline
        \textbf{Explained Variance}                 & 0.946  & 0.933	\\
        \textbf{Mean Squared Error}                 & 0.005  & 0.007	\\
        \textbf{Mean Error}                         & 0.069  & 0.079	\\
        \textbf{Maximum Error}                      & 0.187  & 0.211	\\
        \textbf{Average Absolute Error}             & 0.051  & 0.062	\\
        \textbf{Coef. of determination ($R^2$)}     & 0.943  & 0.922	\\
        \textbf{Pearson's Correlation Coefficient}  & 0.973  & 0.971	\\ \hline
        \end{tabular}
\label{tab:res_45}
\end{table}

\begin{figure}[!ht]
    \centering
    \includegraphics[width=0.9\textwidth]{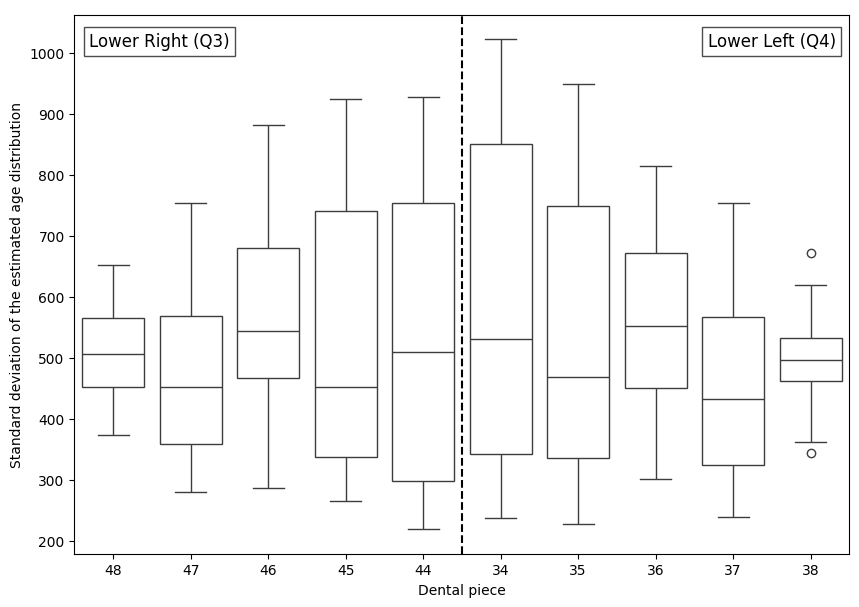}
    \caption{Distribution of the computed uncertainty in the dental age estimation of each mandibular tooth by the opaque (Vila-Blanco) method. On the left (i.e., the right side of the oral cavity), teeth 48 to 44 are present. Conversely, on the right (left side of the oral cavity), teeth 34 to 38 are present}
\label{fig:uncertainty_distribution}
\end{figure}

Regarding the reliability of estimations from the opaque method, Table~\ref{tab:average_deviation_uncertainty} presents the mean and standard deviation of the quantified uncertainty for the clinical cases under study. The complete normal distribution is illustrated in Fig.~\ref{fig:uncertainty_distribution}. This shows that teeth of the same classification (e.g., premolars, molars, and third molars) and specifically symmetrical pieces exhibit analogous uncertainty distributions. This phenomenon can be attributed to the fact that these teeth undergo analogous developmental processes, which leads to the manifestation of comparable characteristics:
\begin{table}[!ht]
\centering
\caption{Mean and standard deviation (STD) of the uncertainties ($\bm{\sigma_t}$) of the dental age estimation (in days) made by the opaque method.}
\begin{tabular}{ccr}
    \hline
    \textbf{Tooth} & \textbf{Mean} & \multicolumn{1}{c}{\textbf{STD}} \\ \hline
    \textbf{48}           & 508.965     &  70.659    \\
    \textbf{47}           & 467.895     & 122.504   \\
    \textbf{46}           & 568.075     & 125.822   \\
    \textbf{45}           & 532.535     & 209.011   \\
    \textbf{44}           & 536.188     & 234.078   \\ \hline
    \textbf{34}           & 586.497     & 254.705   \\ 
    \textbf{35}           & 542.237     & 214.401   \\
    \textbf{36}           & 558.984     & 128.960   \\
    \textbf{37}           & 450.168     & 132.235   \\
    \textbf{38}           & 500.103     &  61.807    \\ \hline
\end{tabular}
\label{tab:average_deviation_uncertainty}
\end{table}
\begin{itemize}
    \item Teeth 34, 35, 44, and 45 (mandibular premolars) exhibit a low average uncertainty. However, these teeth exhibit a significantly greater deviation from the mean compared to other teeth. Notably, permanent premolars exhibit high predictability due to their developmental timeline, which occurs within the age range of two to $14$ years. This characteristic enhances the reliability of age estimation for these teeth.
    \item Teeth 36, 37, 46, and 47 (mandibular first and second molars) have a moderate average uncertainty. In contrast, the deviation from the mean is smaller than in the premolars. The development of the first and second molars is analogous to that of the premolars. However, it is important to note that the reliability of age estimation for these teeth is comparatively lower.
    \item Teeth 38 and 48 (third mandibular molars, i.e., wisdom teeth) teeth exhibit a high average uncertainty and a small deviation. Although both teeth are classified as molars, specifically third molars, their development is irregular. The formation process typically commences at $8$ and concludes at $25$. Consequently, the reliability of these teeth is low.
\end{itemize} 

In summary, the analysis of the uncertainty in the age estimations for the different dental pieces reveals clear patterns related to their classification and developmental timelines. These findings underscore the importance of considering the specific developmental characteristics and age ranges of each tooth type when applying age estimation methods. The variation in uncertainty across different teeth points to potential areas where the model could be refined, particularly for third molars, whose unpredictable development significantly impacts the precision of their age predictions. 

In this study, determining the reliability of opaque method estimations has been considered of great importance. The aggregated uncertainty of the dental age estimation will be calculated using the quantification model proposed by Zadeh. Therefore, it is necessary to obtain the arithmetic mean and standard deviation of the uncertainty determined by the Vila-Blanco method for the dental pieces. The obtained measurements will be employed to establish thresholds that allow determining the degree of uncertainty present in a specific case in relation to the rest of the recorded cases.

\subsection{Human expert validation by dental professional experts}
\label{sec:q1}
In this section, we describe the double validation procedure we followed to assess the quality of the automatically generated explanations. On the one hand, and following the best practices in the field, the quality of the generated explanations was manually validated by dental experts using a questionnaire in which five dimensions were considered. We first describe the collected data in response to the questionnaire, which was designed to validate the automated narratives (Section~\ref{sec:q11}). On the other hand, we also performed a trustworthy self-assessment procedure following the ALTAI checklist, where we considered the seven dimensions of the AI Trustworthiness Assessment List (Section~\ref{sec:q12}).

\subsubsection{Questionnaire Responses}
\label{sec:q11}
The NLG module of AgeX was utilized to generate the textual descriptions associated with four clinical cases corresponding to patients with a chronological age between $5$ and $16$. 
For instance, the following is an excerpt from the fourth ID:
\begin{itemize}
    \item The real age is 7 years and 2 months.
    \item The age estimated by the opaque method is 7 years and 9 months.
\end{itemize}

Seven dental experts were surveyed using a questionnaire to evaluate the quality of four case reports, with descriptions automatically generated by AgeX. 
The results of this survey are summarized in Table \ref{tab:score}, which presents the mean score for each question for all scenarios. 
\begin{table}[!ht]
    \centering
    \caption{Mean and standard deviation (STD) of the score given by a group of dental experts to the three scenarios under study. In Scenario 1 (S1), assessors are provided solely with the dental age estimation. In Scenario 2 (S2), the estimation is accompanied by a concise automated description. In Scenario 3 (S3), the estimation is accompanied by a comprehensive, detailed automated description.}
    \begin{tabular}{lccc}
    \hline
    \textbf{Statement}                                                                                                                                & \textbf{S1} & \textbf{S2} & \textbf{S3} \\ \hline
    \begin{tabular}[c]{@{}l@{}}The textual description uses the appropriate \\ terminology to describe the dentition of the patients.\end{tabular} & 4.57            & 4.54            & 4.64            \\
    The description is easy to understand.                                                                                                         & 4.71            & 4.69            & 4.60            \\
    \begin{tabular}[c]{@{}l@{}}The textual description provided contains enough information.\end{tabular}                                          & 2.54            & 3.50            & 4.89            \\
    \begin{tabular}[c]{@{}l@{}}The textual description provided contains useful information.\end{tabular}                                          & 3.42            & 4.07            & 4.82            \\
    \begin{tabular}[c]{@{}l@{}}The textual description is consistent with the \\ accompanying image (OPG or heat maps).\end{tabular}               & 4.43            & 4.61            & 4.89            \\ \hline
    \multicolumn{1}{r}{\textbf{Mean}}                                                                                                              & 3.94            & 4.28            & 4.77            \\ \hline
    \multicolumn{1}{r}{\textbf{STD}}                                                                                                               & 0.83            & 0.44            & 0.12            \\ \hline
    \end{tabular}
    \label{tab:score}
\end{table}




In the first scenario, assessors were provided solely with concise contextual information and the dental age estimation. 
The mean score in this scenario was the lowest of the three scenarios under study ($3.94$). However, the assessors' perceptions of the adequacy of the terminology ($4.57$), comprehensibility ($4.71$), and consistency with the provided OPG ($4.43$) were favorable. However, the primary concern expressed by assessors pertains to the perceived insufficiency of the information provided ($2.54$) and its perceived lack of utility ($3.42$). This scenario exhibits an STD of $0.83$, signifying the most pronounced divergence among experts of the three scenarios. Furthermore, a mere $3.57$\% of the assessors selected the first scenario as the optimal one. 

In the second scenario, assessors receive a brief description accompanied by the OPG. The mean score attained was ($4.28$). The evaluators concur that the quality of the narrative is commendable, as the average score assigned to this scenario exceeds $4$. However, they perceive the information provided with the brief descriptions as insufficient ($3.5$), though it is superior to the narrative of the first scenario. The STD for this scenario is $0.44$. It is noteworthy that none of the assessors selected the second scenario as the optimal choice.

In the third scenario, assessors are provided with a detailed description and a heat-map. The highest average score is reached at $4.77$. The assessors reached a consensus that the provided narratives employed suitable terminology ($4.64$), were straightforward ($4.60$), furnished adequate ($4.89$) and pertinent ($4.82$) information, and were congruent with the accompanying images ($4.89$). Furthermore, the STD of the ratings obtained in this scenario is lower than in the other two scenarios ($0.12$ vs. $0.83$ and $0.44$), indicating a higher degree of agreement among experts and, consequently, lower uncertainty. Furthermore, $89.29$\% of the assessors selected the third scenario as the optimal one.

In sum, the prevailing sentiment among assessors is a marked preference for the utilization of the comprehensive narratives produced by the NLG module of AgeX as textual support, thereby substantiating the efficacy of the implemented system.

\subsubsection{ALTAI Self-Assessment}
\label{sec:q12}

\begin{figure}[!ht]
    \centering
\includegraphics[width=\textwidth]{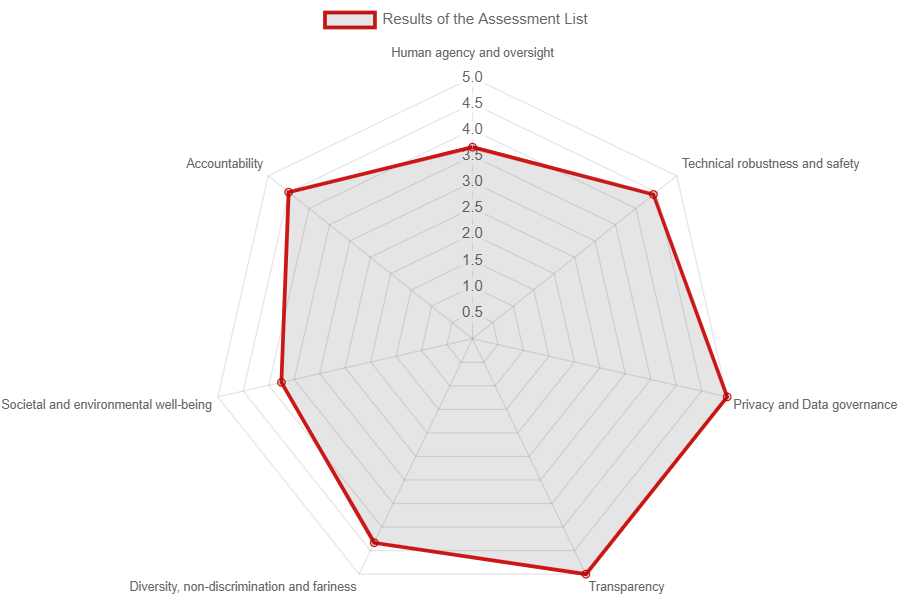}
    \caption{Assessment of trustworthiness, reliability and ethical integrity of the AgeX system according to the 7 ALTAI dimensions: (1) Human agency and oversight, (2) Technical robustness and safety, (3) Privacy and Data Governance, (4) Transparency, (5) Diversity, non-discrimination, and fairness, (6) Societal and environmental well-being and (7) Accountability}
    \label{fig:altai}
\end{figure}

The trustworthiness of the AgeX system was evaluated using the ALTAI checklist, with the results shown in Fig.~\ref{fig:altai}. In light of this feedback, we have reflected on how to improve AgeX. The dimensions that received the lowest scores were Human agency and oversight ($3.66$) and Societal and environmental well-being ($3.75$).

On the one hand, with respect to Human agency and oversight, it should be noted that AgeX does not make any autonomous decisions; on the contrary, it is a tool aimed at assisting dental experts who are responsible for decision-making. In accordance with this objective, a disclaimer has been incorporated into the software's operation to explicitly affirm our respect for human autonomy. 

Conversely, the Societal and environmental well-being considerations extend beyond individual users to encompass broader social and ecological impacts. It is imperative to methodically assess the dual impacts of AI systems on the environment, incorporating mechanisms to evaluate and mitigate these effects in a perpetual manner. In addition, there is a compelling rationale for the proactive reduction of the environmental footprint of AI throughout its lifecycle. Furthermore, there is a strong impetus for participation in initiatives or competitions that promote environmentally sustainable AI innovations. As our proposal utilizes a rule-based natural language generator along with a predefined linear regressor for computations, we consider its environmental impact negligible. However, the provision of training opportunities and educational materials to enable users to acquire new skills or improve existing ones remains a crucial aspect that requires further consideration. 

In conclusion, with regard to the remaining ALTAI dimensions, it is deemed that no additional modifications are required. This is due to the fact that AgeX is a prototype intended to validate our research hypothesis and it will be utilized exclusively for dissemination purposes.

\section{Ethical and Legal Issues}
\label{sec:els}
The work described in previous sections were approved by the Research Ethics Committee of the University of Santiago de Compostela (USC). 
The committee has authorized ``extracting relevant characteristics of panoramic dental radiographs for several purposes, among which are the prediction of chronological age, the detection of wounds, the diagnosis of oral diseases, or the monitoring of clinical processes". The permit stipulates that research in these domains with the dataset containing the OPGs is under the ethical guidelines established by USC. 
Consequently, the image collection procedure was executed with the approval of the USC Research Ethics Committee. 
Additionally, an anonymization process was meticulously executed, entailing the removal of any identifiable subject information from the image metadata. This was a prerequisite for obtaining the committee's approval. Consequently, the data retained for each image consists exclusively of the essential details necessary for dental age estimation, thereby ensuring that patient identification is rendered unfeasible. 

In developing and implementing AI systems, it is imperative to consider ethical considerations to ensure that the technology benefits society in a fair, transparent, and responsible manner~\citep{ethics_guidelines_2019}. As AI continues to be integrated into various domains, it is essential to evaluate its impact not only in terms of performance but also through an ethical lens. This analysis is instrumental in ensuring that AI systems are developed in a manner that respects human rights, fosters fairness, and circumvents potential harm. 
Addressing these ethical aspects, as previously outlined in the ALTAI self-assessment, enables a comprehensive evaluation of AgeX's alignment with established ethical principles and guidelines, thereby ensuring its trustworthiness for deployment in real-world applications. 

Regarding transparency, we introduced an interpretable surrogate model. This allows us to explain the automated decisions in terms that are easier to grasp so dental experts (i.e., our end-users) can better understand the inner workings of the dental age estimation. The surrogate approximates the decision-making process, making it possible to trace how inputs lead to specific outcomes. 
Moreover, the surrogate helps generate explanations by providing insight into which features influence automated decisions. These explanations are presented in human-friendly terms, assisting users in understanding the rationale behind estimations.

In consideration of the principles of diversity, non-discrimination, and fairness, it is acknowledged that disparate populations may manifest natural variations in tooth development due to genetic factors, nutritional influences, and environmental determinants. These variations are not associated with demographic characteristics (e.g., race, gender, or socioeconomic status) but reflect biological differences across populations.
With respect to the variable of sex, it is imperative to note that all instances of data are balanced against each other across all age ranges. However, it is acknowledged that potential biases may be present in the data due to the inclusion of patients from the USC Faculty of Medicine and Dentistry of USC. Patients of all ages are included, but according to the dental expert who collected the data, most of them are aware or suspect that they have some form of oral pathology. Consequently, the proportion of subjects in the dataset who are free of oral pathologies is comparatively low.

In addition, the following characteristics of the data reveal additional biases:
\begin{itemize}
    \item Most of the data are from patients with special oral health considerations or unique dental anatomy. For example, there are some OPGs from patients with missing teeth.

    \item In general, the people with dental problems who need to use research centers have the lowest incomes because they generally cannot afford dental treatment.

    \item All the data were collected in Santiago de Compostela, an area with a majority of Caucasian population.
\end{itemize}


Legally, the data processing in AgeX is done in compliance with the European GDPR, as we work with sensitive personal data (i.e., biometric information that could be processed to identify a person) in a high-risk medical application. Specifically, we have paid particular attention to GDPR Article 9, Paragraph 1: ``Processing of personal data revealing racial or ethnic origin, political opinions, religious or philosophical beliefs, or trade union membership, and the processing of genetic data, biometric data to uniquely identify a natural person, data concerning health or data concerning a natural person’s sex life or sexual orientation shall be prohibited.''

It should be noted that this prohibition does not apply if one of the exceptions mentioned in Article 9 is met. 
In particular, exception ``H'' specifies that: ``Processing is necessary for the purposes of preventive or occupational medicine, for the assessment of the working capacity of the employee, medical diagnosis, the provision of health or social care or treatment, or the management of health or social care systems and services on the basis of Union or Member State law or pursuant to a contract with a health professional and subject to the conditions and safeguards referred to in paragraph 3.''

In addition, paragraph 3 aforementioned states that data processing must be carried out by a qualified professional and is subject to professional secrecy under the EU or national law. Accordingly, AgeX may only be used by skilled professionals.

The AI Act classifies AI healthcare applications as high-risk and emphasizes in this case the necessity for these systems to be transparent, providing users with clear and understandable information about their operation and limitations. In our case, AgeX places a strong emphasis on ensuring transparency. Accordingly, AgeX has been carefully designed to be transparent and self-explanatory through natural narratives.

\section{Conclusions and Future Work}
\label{conc}
Cameriere's manual method is a well-known transparent method used in dentistry for estimating dental age. 
In contrast, the automatic deep learning model proposed by Vila-Blanco et al. is an opaque, untraceable model. 
In both methods, the interpretation of the relevance of the variables (linear regression coefficients in one case, heat-map colors in the other) is left to the system users. In this work, we have explored how to increase trustworthiness by providing AgeX with self-explanatory capabilities. 

Accordingly, the primary objectives of this study were: (i) to analyze the opaque method based on deep learning techniques (automatic method) and compare it with the transparent method based on linear regression techniques (manual method); (ii) to determine whether explanations could be generated for the dental age estimates made by the automatic method using the manual method taken as a surrogate method; (iii) to construct understandable narratives for the experts that allow them to understand how the automatic dental age estimation is performed. 

All objectives were successfully met. 
Specifically, narratives were generated to substantiate the dental age estimation made by the opaque method, with the transparent method serving as a surrogate. To this end, a statistical analysis was performed on the relationship between the two methods. This analysis concluded that both methods exhibited a high degree of correlation, thereby indicating that they yielded equivalent information. A further study assessed the reliability of the estimations to be incorporated into the aforementioned narratives. In addition, a rule-based D2T system was developed to generate these narratives in natural language automatically. The sentences were designed following the specifications of the dentist researcher who co-authored this work. Consequently, the sentences are comprehensible and align with the terminology utilized by experts in the field. The narratives were then subjected to validation by a group of experts. The quality of the narratives was assessed through the administration of questionnaires. Finally, the AgeX open-source software was implemented as a proof of concept, facilitating the execution of the entire experiment in a centralized and straightforward manner.

Finally, it is important to acknowledge that there remains potential for improvement in future research. Namely, we plan to convene a board of experts to redesign the sentences generated by AgeX and to extend the human evaluation in clinical settings. In addition, we will explore new methods for generating multilingual explanations.

\bigskip





\begin{appendices}

\section{Original Generated Texts (Spanish, Patient\#2324)}\label{app1:original_texts}
\subsection{Introduction}
\begin{displayquote}
Una niña con una edad cronológica real de 7 años y 2 meses debería presentar dentición mixta.
\end{displayquote}

\subsection{Short description}
\begin{displayquote}
La edad cronológica estimada es de 7 años y 9 meses porque no tiene ningún diente completamente formado en el cuadrante mandibular derecho y el ápice del segundo premolar mandibular derecho está bastante abierto.
\end{displayquote}

\subsection{Detailed description}
\begin{displayquote}
La edad cronológica estimada, de acuerdo con el método de Vila-Blanco et al., es de 7 años y 9 meses porque no tiene ningún diente completamente formado (0 de 5) en el cuadrante mandibular derecho y el ápice del segundo premolar mandibular derecho (45) está bastante abierto.

La edad cronológica real es 7 años y 2 meses. Por tanto, existe una sobreestimación (7 meses) con respecto a la edad real. La fiabilidad de esta estimación es alta, siendo las piezas dentales más predictivas la 44 y la 37

La estimación de la edad del paciente se ha realizado utilizando el método de Vila-Blanco et al., una aproximación automática basada en aprendizaje profundo. Esta metodología detecta los dientes del paciente en su OPG y calcula las distribuciones probabilísticas asociadas a la edad cronológica de cada diente. La edad del paciente se estima mediante la agregación de las edades de sus piezas dentales. Con la información inferida por el método automático se puede construir un mapa de calor sobre la OPG correspondiente, tal y como se ve en la figura mostrada. Para ello, se le otorga un color a cada píxel de cada diente mandibular dependiendo de su importancia para la estimación de la edad cronológica.
\end{displayquote}

\end{appendices}

\newpage

\section*{Ethics approval and consent to participate}
This study was conducted in accordance with ethical standards and approved by the Ethics Committee of the University of Santiago de Compostela. 

\section*{Consent for publication}
All authors have reviewed the final manuscript and consent to its publication.

\section*{Availability of data and material}
The data and code used in this study are not publicly available due to privacy and institutional restrictions. However, they may be made available upon reasonable request, and such requests will be considered on a case-by-case basis.

\section*{Competing interests}
The authors declare that they have no known financial or personal conflicts of interest that could have appeared to influence the work reported in this paper.

\section*{Funding}
\sloppy Ainhoa Vivel Couso is a PhD Researcher supported by the Galician Ministry of Education, Science, Universities and Professional Training (ED481A-2024-059). All authors recognize the support of the Galician Ministry of Education, Science, Universities and Professional Training (CiTIUS 2024-2027 ED431G2023/04, GSI 2022-2025 ED431C2022/19, GVA GRC2021/48 and OSRG GPC2020/27 grants) and the European Union (European Regional Development Fund - ERDF). This work is also supported by the Spanish Ministry of Science, Innovation and Universities (MCIN/AEI/10.13039/501100011033/) through XAI4SOC PID2021-123152OB-C21 and COLLAGE PID2023-149549NB-I00 grants and the Instituto de Salud Carlos III (Spain) through PI21/00588. 

\section*{Author contributions}
Author contributions are provided according to the Contributor Roles Taxonomy (CRediT): 
\begin{itemize}
    \item Ainhoa Vivel-Couso: conceptualization (supporting), data curation (lead), formal analysis (lead), funding acquisition (equal), investigation (lead), methodology (equal), software (lead), validation (lead), visualization (lead), writing – original draft (lead), writing – review \& editing (supporting). 
    \item Nicolás Vila-Blanco: conceptualization (equal), data curation (supporting), formal analysis (supporting), investigation (supporting), methodology (equal), resources (lead), supervision (lead), writing – original draft (supporting), writing – review \& editing (equal). 
    \item María J. Carreira: conceptualization (lead), investigation (supporting), methodology (supporting), project administration (equal), resources (equal), supervision (equal), validation (equal), writing – review \& editing (equal). 
    \item Alberto Bugarín-Diz: conceptualization (equal), investigation (supporting), methodology (equal), project administration (equal), resources (equal), supervision (equal), writing – review \& editing (equal). 
    \item Inmaculada Tomás: conceptualization (equal), investigation (supporting), methodology (equal), resources (equal), writing – review \& editing (supporting). 
    \item Jose M. Alonso-Moral: conceptualization (equal), funding acquisition (equal), investigation (equal), methodology (equal), project administration (equal), resources (equal), supervision (equal), validation (equal), writing – review \& editing (lead). 
\end{itemize}


In addition to the contributions listed above, we would like to express our gratitude to those who supported this work. We thank dental expert Paulina Varas-Quintana, DDS, for her assistance during the early design process of the texts generated by AgeX. We also thank Pablo Canosa-García for his contributions to Section~\ref{sec:els} (Ethical and Legal Issues). Finally, we are grateful to all the dental experts who participated in the human evaluation.


\bibliographystyle{apalike}
\bibliography{bib}

\end{document}